\documentclass[runningheads]{llncs}

\usepackage[T1]{fontenc}
\usepackage{graphicx}
\usepackage{multirow}
\usepackage{float}
\usepackage{amsmath}

\title{\textit{ToDER}: Towards Colonoscopy Depth Estimation and Reconstruction with Geometry Constraint Adaptation}

\author{Zhenhua Wu*\inst{1} \and
Yanlin Jin*\inst{2} \and
Liangdong Qiu*\inst{3} \and
Xiaoguang Han\inst{3} \and
Xiang Wan\inst{3} \and
Guanbin Li\textsuperscript{\dag}\inst{1}}

\institute{
\inst{} Sun Yat-sen University \and
\inst{} Rice University \and
\inst{} Chinese University of Hong Kong (Shenzhen) \\[1ex]
\textbf{Emails:} 
\email{wuzhh56@mail2.sysu.edu.cn}, 
\email{Yanlin.Jin@rice.edu}, 
\email{liangdongqiu@link.cuhk.edu.cn}, 
\email{hanxiaoguang@cuhk.edu.cn}, 
\email{wanxiang@sribd.cn}, 
\email{liguanbin@mail.sysu.edu.cn}
}

\begin{document}

\maketitle       

\begin{abstract}
Visualizing colonoscopy is crucial for medical auxiliary diagnosis to prevent undetected polyps in areas that are not fully observed. Traditional feature-based and depth-based reconstruction approaches usually end up with undesirable results due to incorrect point matching or imprecise depth estimation in realistic colonoscopy videos. Modern deep-based methods often require a sufficient number of ground truth samples, which are generally hard to obtain in optical colonoscopy. To address this issue, self-supervised and domain adaptation methods have been explored. However, these methods neglect geometry constraints and exhibit lower accuracy in predicting detailed depth. We thus propose a novel reconstruction pipeline with a bi-directional adaptation architecture named \textit{ToDER} to get precise depth estimations.   %We thus propose a novel reconstruction pipeline with a bi-directional adaptation architecture. We first propose a depth estimation method \textit{ToDER} to get depth maps. 
Furthermore, we carefully design a TNet module in our adaptation architecture to yield geometry constraints and obtain better depth quality. Estimated depth is finally utilized to reconstruct a reliable colon model for visualization. Experimental results demonstrate that our approach can precisely predict depth maps in both realistic and synthetic colonoscopy videos compared with other self-supervised and domain adaptation methods. Our method on realistic colonoscopy also shows the great potential for visualizing unobserved regions and preventing misdiagnoses.
\end{abstract}
\renewcommand{\thefootnote}{}
\footnotetext{* Equal contribution.}
\footnotetext{\textsuperscript{\dag} Corresponding author.}

\section{Introduction}
Colorectal cancer is the most lethal malignancies globally while colonoscopy is currently the most direct and effective method for detecting intestinal lesions such as polyps and colon cancer. %During colonoscopy, if any significant lesions are encountered, specific procedures could be performed to prevent further potential disease. Lesions on uninvestigated regions caused by lack of camera orientation or occlusion on colon structures would be dangerous. 
Reducing the percentage of colon surfaces missed during colonoscopy is necessary for polyps detection~\cite{hong2007colonoscopy}. Therefore, we need an effective reconstruction method for colonoscopy videos to illustrate the detected colon surface and uninvestigated area. Such an approach could serve as a visual cue for healthcare practitioners to revisit the unexamined regions, thereby minimizing the risk in disease detection. 

\begin{figure}[!t]
	\centering
	\includegraphics[width=0.9\textwidth]{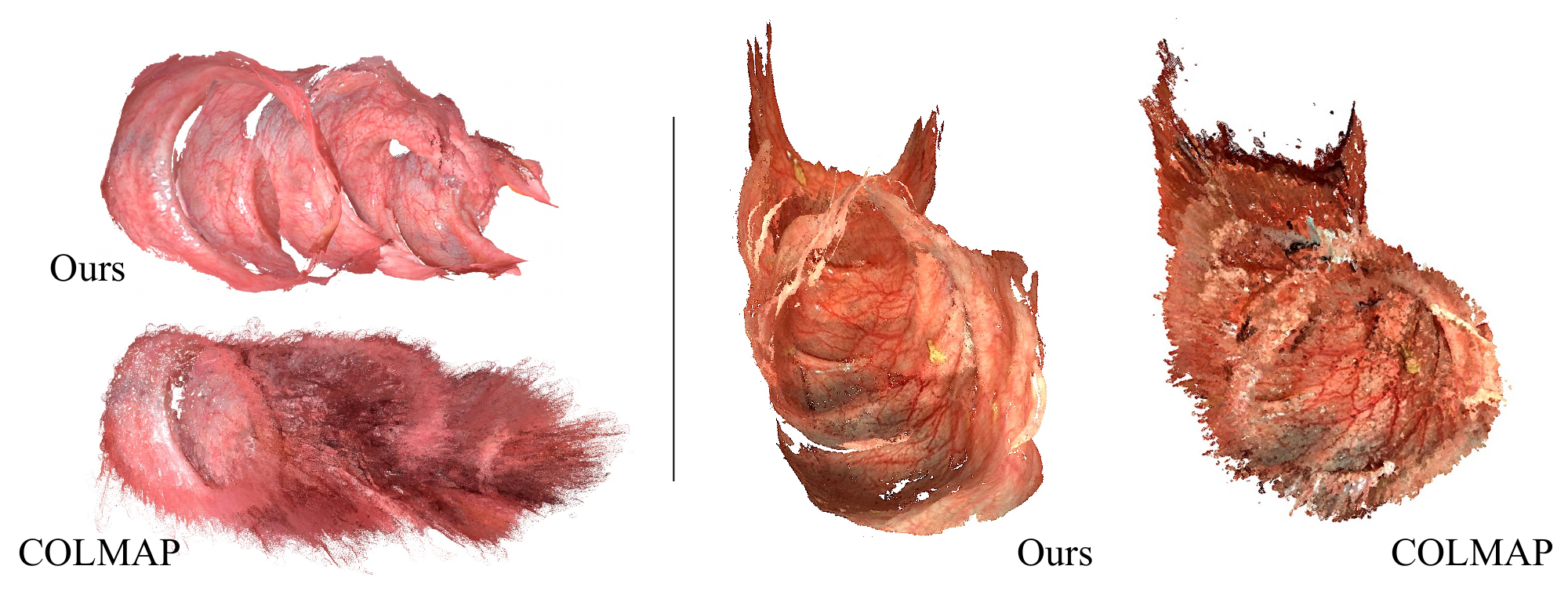}
 	\vspace{-4mm}
	\caption{\textbf{Colon models reconstructed from two realistic colonoscopy videos}. We compared our method with COLMAP~\cite{schonberger2016structure}. It shows that our method is able to reconstruct high-quality colon surface and clear unobserved area for reconstruction.}
	\label{fig:teaser}
 	\vspace{-4mm}
\end{figure}

Traditional reconstruction methods often suffer suboptimal performance in colonoscopy images. They demand matched image features to recover geometry knowledge. However, in colonoscopy images, limited number of discernible key points make this difficult. Furthermore, the instability in lighting and reflections, caused by camera movements and intestinal peristalsis, often leads to failed reconstructions. %Modern deep learning techniques for colonoscopy~\cite{ma2021rnnslam,zhang2021colde,chen2019slam} commonly involve the creation of a 3D colon model using optical colonoscopy (OC) data. 
Modern deep-based reconstruction method often predict depth maps and camera poses to complete 3D models. However, colonoscopy depth estimation is extremely challenging. These depth estimation methods have made promising progress in scenarios such as outdoor driving~\cite{cordts2016cityscapes} and scene reconstruction~\cite{li2018megadepth} due to the availability of labelled depth in the supervised setting. To address the challenge where ground truth depth maps are hard to obtain, self-supervised learning based techniques~\cite{bian2021unsupervised,godard2019digging,zhou2017unsupervised} and domain adaptation methods~\cite{zheng2018t2net,chen2021s2r} offer viable solutions. In the context of colonoscopy reconstruction, self-supervised approaches are employed for depth maps generation in realistic video sequences~\cite{zhang2021colde,posner2022c}. For domain adaptation, Mahmood \textit{et al.}~\cite{mahmood2018unsupervised} transforms the realistic images into synthetic domain and then predicts depth with a network pretrained on synthetic data. Rau \textit{et al.}~\cite{rau2019implicit} develop a generative model on synthetic data showing implicit domain adaptation. 

Ensuring consistency is important in colonoscopy reconstruction for both feature-based traditional methods and deep-based methods. Traditional methods rely on optimization techniques that require consistent and paired features~\cite{schonberger2016structure,mur2015orb}. Similarly, deep-based SLAM reconstruction approaches~\cite{ma2021rnnslam,chen2019slam,posner2022c} employ predictive models to establish consistent camera poses. The scale consistency of depth maps, a crucial factor for accurate reconstruction, can be effectively maintained through self-supervised techniques~\cite{godard2019digging,bian2019unsupervised,bian2021unsupervised}. However, it is noteworthy that existing domain adaptation methods~\cite{mahmood2018unsupervised,rau2019implicit} have predominantly focused on domain translation for only the image dataset. They neglect the preservation of geometry constraints between adjacent frames, result in region misalignment and introduce potential noise during the reconstruction process.

To address above issues, we present a novel method for optical colonoscopy reconstruction. We first propose a bi-directional domain adaptation structure with our designed TNet. Our TNet predict the pose for geometry constraints. Specifically, these constraints encourage inter-frame photometric continuity, penalize variations between adjacent depth, and greatly improves the depth results. 
A multi-stage training strategy is employed to gradually get precise depth maps for colonoscopy images. Finally, our fused depth results, combined with surfel-based reconstruction methods~\cite{schonberger2016structure,schops2019surfelmeshing}, facilitate the 3D reconstruction of colon surfaces.

To summarize, our contribution can be listed as follows:
\begin{itemize}
    \item We introduce a novel architectural framework for colonoscopy 3D reconstruction using geometry constraint adaptation and obtain accurate depths on colonoscopy videos. 
    \item Our depth estimation approach incorporates an innovative TNet module capturing geometric information between consecutive frames, which enhances the depth accuracy and consistency. 
    \item  Our experiments demonstrate the superiority of \textit{ToDER} compared with other self-supervised and domain adaptation methods qualitatively and quantitatively and yield fine reconstruction results on realistic colonoscopy.
\end{itemize}

\begin{figure*}[t]
	\centering
	\includegraphics[width=0.98\textwidth]{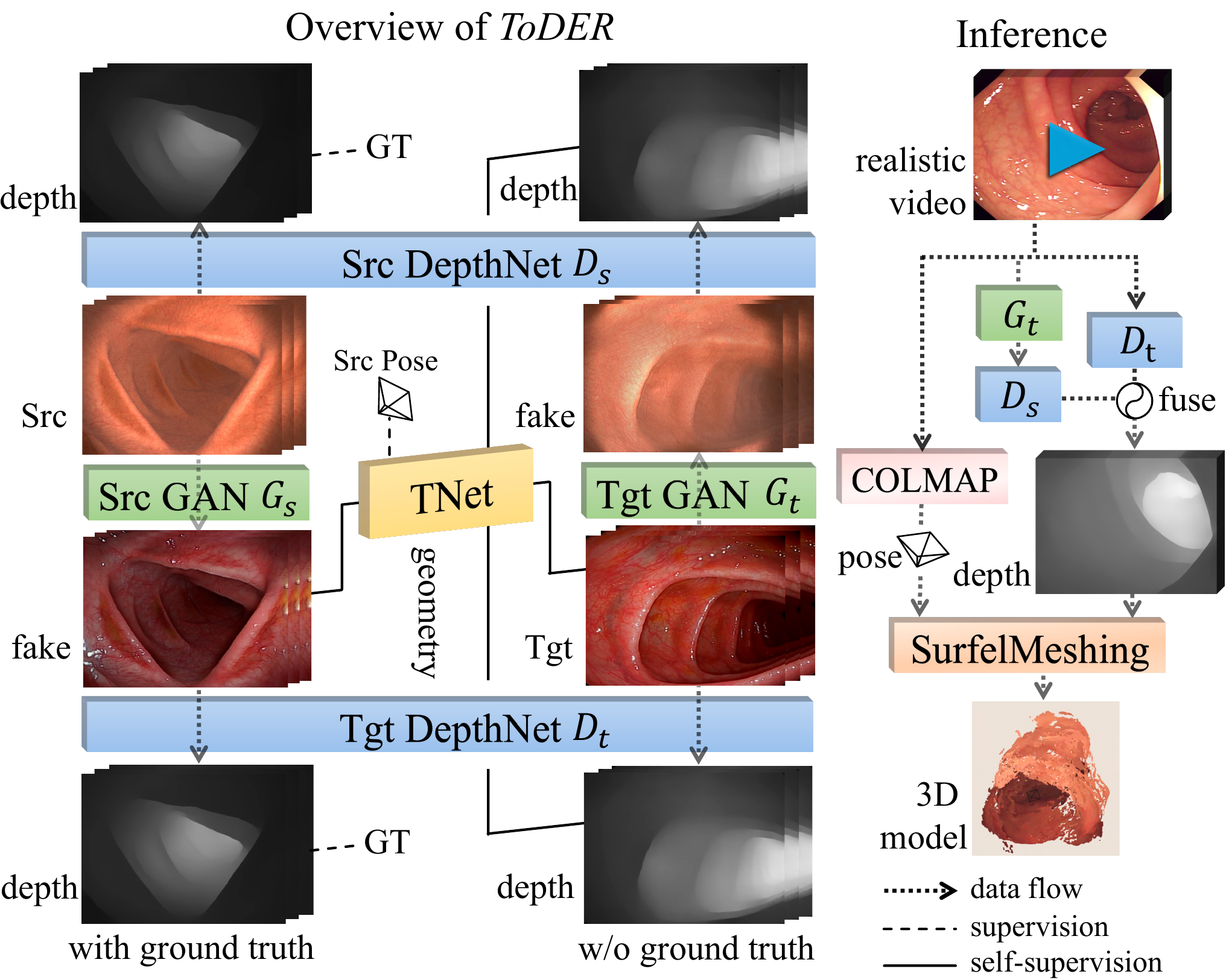}
 	\vspace{-4mm}
	\caption{\textbf{Pipeline} of our method. The \textbf{left part} denotes the overview of \textit{ToDER} for depth estimation. Starting from Src and Tgt (which indicating the source and target domain), the GANs are first initialized to transfer surface styles between two domains. DepthNets and TNet are next trained and fine-tuned to predict depths and improve the depth quality respectively. All the modules are supervised through the corresponding guidance denoted by the lines. The \textbf{right part} illustrates the reconstruction process using \textit{ToDER} to get a reconstructed colon model from realistic or synthetic videos. Two depth maps predicted from the original target style and converted source style are next fused to infer the result depth. A 3D model is finally built by SurfelMeshing given pose from COLMAP and depth from \textit{ToDER}. Note that \textit{ToDER} does not need to label the depth on realistic colonoscopy.}
	\label{fig:pipeline}
 	\vspace{-4mm}
\end{figure*}

\section{Method}

\subsubsection{Overview}
We devise a novel domain adaptation method \textit{ToDER} to acquire a robust depth estimation on colonoscopy videos. Inspired by~\cite{zhu2017unpaired}, our pipeline applies a bi-directional style. As shown in the left part of Fig~\ref{fig:pipeline}, we illustrate Depth estimation training process for \textit{ToDER}. The two columns from left to right, represent the source-to-target and target-to-source data flow. The generative models $G_s \& G_t$ are used to convert the image style into the other domain and $D_s \& D_t$ to predict depth. TNet in the center provides geometric constraints to improve depth estimations. For right part of Fig~\ref{fig:pipeline}, it shows how we adopt ToDER and other techniques to obtain final models.

%and take advantage of all pre-trained $D_s, D_t, G_s$ and $G_t$. The TNet exploits the epipolar geometry information present between consecutive frames and depth estimation, which further strengthens our domain adaptive model and significantly benefits depth estimation.

% in the first step. In the second step, we leverage the generative models $G_s$ and $G_t$ to generate synthetic images, for example, the fake target images $x^i_{s2t}=G_s\left(x_s^i\right)$ from the source domain paired with depth annotation. Two DepthNets $D_s, D_t$ are followed to predict the depth of the source and target domain, respectively. We finally achieve accurate and robust depth estimation performance by using generated outputs from both $G_s$ and $G_t$. Moreover, we introduce a TNet to exploit the epipolar geometry information present between consecutive frames. This further strengthens our domain adaptive model and significantly benefits depth estimation.

\subsubsection{Style Translation}
Given a set of $N$ synthetic source images $x_s^i \in X_s, i=1,\cdots,N $ with paired depth $y_s^i$, we aim to build a depth estimation model for the image $x_t$ in realistic colonoscopy, the target domain. Our built synthetic dataset serves as the source domain. The ground truth depths and camera poses provided in the source domain are utilized in a supervised manner to guide the domain adaptation task. Our objective is to use the ground truth knowledge present in the source domain as a guiding principle to generalize the depth estimation model for realistic scenarios. We initialize the domain style translation by first training $G_s$ and $G_t$ to bridge the gap between the source domain $X_s$ and the target domain $X_t$. 

%Besides the vanilla minimax adversarial loss, identity mapping loss~\cite{taigman2016unsupervised} and cycle consistency loss~\cite{zhu2017unpaired} are also adopted. As a result, the loss for generative models is expressed as:
% \begin{equation}
%  \mathcal{L}_{gan} = \mathcal{L}_{minimax} + \lambda_1 \mathcal{L}_{i d t} + \lambda_2 \mathcal{L}_{c y c} 
% \label{eq:gan}
% \end{equation}

% where $\lambda_1$ and $\lambda_2$ are the weights of $\mathcal{L}_{i d t}$ and $\mathcal{L}_{c y c}$ and are set to $30$ and $1$, respectively.

\subsubsection{Depth Prediction}
Next, we predict depth from source and target domain images. Denote these two DepthNets as $D_s$ and $D_t$. Specifically, for the case of $D_t$, leveraging pre-trained generative models $G_s$, we expand the training dataset $x^i_{s2t}=G_s\left(x_s^i\right)$ by incorporating ground truth depth information $y_s^i$. An $\ell_1$ loss is applied on the pairs $\left(x^i_{s2t},y_s^i\right)$ for supervision, and similarly for $D_s$. Although depth ground truth for the target images predicted by $D_t \& D_s$ is unknown, the predicted depth are expected to be the same since they essentially depict the same target image. They are then optimized in a self-supervised manner.

% A local smoothness loss $\mathcal{L}_{sm}$ for $Y_t$ and $Y_{t2s}$ is introduced to encourage smooth depth transitions in local regions. This loss helps to prevent abrupt changes in depth values and promotes spatial consistency in the estimated depths. We minimize the following loss to initialize the DepthNets:

% \begin{equation}
% \begin{aligned}
% \mathcal{L}_{D_t} &= \alpha_1 \left\|y_s-\tilde{y}_{s2t}\right\| + \alpha_2 \mathcal{L}_{sm}\left(\tilde{y}_{s2t}\right \\
% \mathcal{L}_{D_s} &= \alpha_1 \left\|y_s-\tilde{y}_{s}\right\| + \alpha_2 \mathcal{L}_{sm}\left(\tilde{y}_{s}\right  \\
% \mathcal{L}_{pose} &= \left\|P(X_s)-P(X_{t2s})\right\|
% \end{aligned}
% \label{eq:depth}
% \end{equation}

% where $\alpha_1$ and $\alpha_2$ are the weights of two losses and are set to $1$ and $0.5$. $P(X_s)$ is the relative pose set of $X_s$.

\subsubsection{TNet for Geometric Constraints}
We observe that current self-supervised and domain adaptation methods fail to accurately predict depth details in colonoscopy videos. They focus on the image style consistency but style texture could be unreliable when reflection and intestinal peristalsis exist. To solve this, we proposed TNet for geometry constraints. The geometry serves a stable supervision when style texture provides noisy information. Specifically, TNet learns the transformation between video frames and serves as a rough pose predictor for the input consecutive images. It takes two consecutive frames as input and outputs their relative transformation. Estimated transformation from TNet and depth estimation from pretrained DepthNets allow us to warp one frame onto another. This allows photometric consistency for supervision between two image frames. We use the $L1$ norm and the element-wise similarity SSIM as loss functions:

\begin{equation}
\mathcal{L}_p\left(I\right)=\frac{1}{|I|} \sum_{p \in I}\left(\lambda_i\left\|I(p)-I^{\prime}(p)\right\|_1+\lambda_s \frac{1-\operatorname{SSIM}\left(I, I^{\prime},p\right)}{2}\right) 
\label{eq:tnet_self}
\end{equation}

where $I$ denotes the input reference frame, and $I^{\prime}$ denotes the warped image from the neighboring image of $I$. In our case, we choose $\alpha_3=0.1$.

We further penalize the inconsistency between depth predictions similarly with warping. We measure the correlation between the points in the current frame and the differentiable bi-linear interpolation of the corresponding points in the neighboring frame, with loss expressed as:
\begin{equation}
% \begin{aligned}
%  &\mathcal{S}_{cons}(p) = \frac{|{M}_{a}(p)-{M}_{b}(p)|}{{M}_{a}(p)+{M}_{b}(p)} \\
%  &\mathcal{L}_{cons} = \frac{1}{|V|}\sum_{{p}\in{V}}{S}_{cons}(p)
% \label{eq:frames}
% \end{aligned}
\mathcal{S}_{cons}(p) = \frac{|{M}_{a}(p)-{M}_{b}(p)|}{{M}_{a}(p)+{M}_{b}(p)},   \mathcal{L}_{cons} = \frac{1}{|V|}\sum_{{p}\in{V}}{S}_{cons}(p)
\label{eq:frames}
\end{equation}

where ${M}_{a}(p)$ and ${M}_{b}(p)$ are the corresponding points of depth map. $V$ denotes valid points set successfully projected between the corresponding frames.

\subsubsection{Multi-stage Training} 
There are three stage for training: (1) Two GANs are initialized from synthetic and realistic data. (2) Training depth prediction models $D_s$ and $D_t$. (3) Refine all the parameters with TNet to consolidate geometric information and ensure inter-frame consistency. Please refer to the Supplementary for detailed loss equations.

%The loss function of this step is:
% \begin{equation}
%  \mathcal{L}_{step1} =  \mathcal{L}_{gan} = \mathcal{L}_{minimax} + \lambda_1 \mathcal{L}_{i d t} + \lambda_2 \mathcal{L}_{c y c}
% \label{eq:gan}
% \end{equation}

% In the second step, we perform pre-training of two deep networks $D_s$ and $D_t$, using the colonoscopy data $X_s$ and $X_{s2t}$. The loss function of this step is:
% \begin{equation}
%  \mathcal{L}_{step2} =  \mathcal{L}_{D_t} + {L}_{D_s}
% \label{eq:gan}
% \end{equation}

% Lastly, to consolidate geometric information and ensure interframe consistency, we have incorporated TNet into the two-way domain adaptation training process. The loss function of this step is:
% \begin{equation}
%  \mathcal{L}_{step3} =  \mathcal{L}_{gan} + {L}_{D_t} + {L}_{D_s} + {L}_{pose} + {L}_p\left(I\right) + {L}_{cons}
% \label{eq:gan}
% \end{equation}

\subsubsection{Reconstruction} 
Once we have obtained the depth and camera poses from COLMAP, we employ a  sufel-based reconstruction~\cite{schops2019surfelmeshing} approach to construct a 3D mesh representation of the colon since the surfels carry more information compared with point cloud and Poisson reconstruction in COLMAP.

\begin{figure*}[ht]
	\centering
	\includegraphics[width=0.98\textwidth]{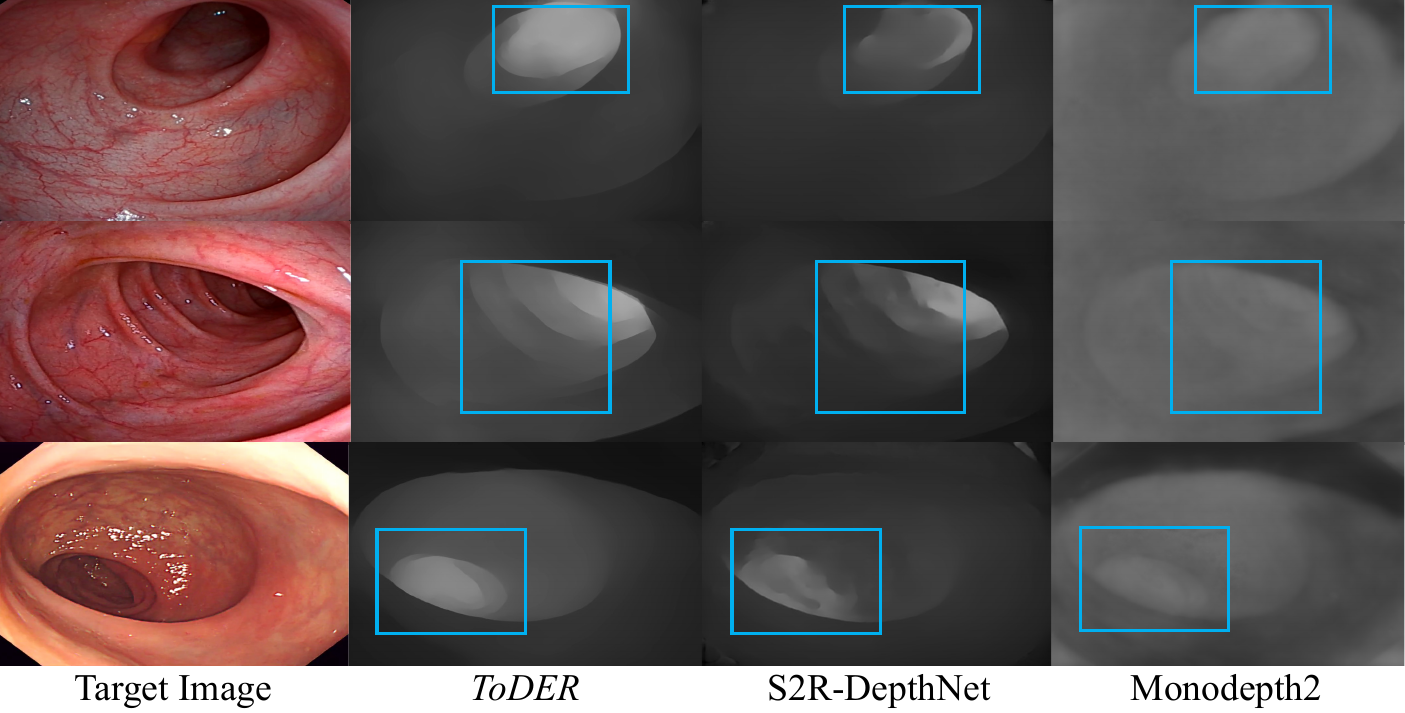}
 	\vspace{-4mm}
	\caption{\textbf{Depth estimation results on realistic colonoscopy videos.} Although the pose are unknown on realistic video, \textit{ToDER} still produce reasonable depths with clear details while other methods suffer vague depth prediction on details. We show more depth results with compared methods in Fig.~\ref{fig:depth_more}.}
	\label{fig:depth_real}
 	\vspace{-0mm}
\end{figure*}
\section{Experiments}

\subsection{Experiments details}
\subsubsection{Datasets}

We created a large synthetic colonoscopy dataset with two texture styles. The two subsets use different 3D intestinal mesh models derived from CT scans~\cite{zhang2020template,incetan2021vr}. The source virtual dataset (Style A) features a brown albedo map with reduced smoothness (0.38) and subtle vascular structures, while the target one (Style B) has pink hue, smoother texture (0.83), and prominent vessels. Synthetic dataset is built in Unity3D with an RGB camera, a depth camera, a point light, and colon models with our costumed textures. Attached sampling scripts automatically recorded RGB images, depth images (16-bit), poses, and timestamps, formatted according to TUM~\cite{sturm12iros} and support lens distortion, motion blur, and vignetting. For each synthetic style, there are 3,000 training images and four test sets with 200 images each. Additionally, we extracted 3,000 images from realistic colonoscopy videos, using Style A as the source and realistic colonoscopy as the target domain, demonstrating real-world use of our method.
\label{subsec:datasets}

\subsubsection{Implementation} 

Our depth prediction network employs the UNet architecture~\cite{Zheng_2018_ECCV}. For TNet, which handles two sequential RGB frames at a time, we utilize the ResNet-18 encoder and a pose decoder following the posenet of Monodepth2~\cite{godard2019digging}. In order to input two consecutive frames, the input channel number for ResNet is changed to 6, and the final output are 6 DOF relative poses. Training are conducted on a single A100 GPU, optimizing models taking 640 by 480 images through Adam~\cite{kingma2014adam}. The number of epochs and lr in three training stages are 200, 110 \& 110 and $5e^{-5}$, $1e^{-4}$ \& $1e^{-4}$ respectively.

% \subsection{Camera poses and intrinsics}
% In both virtual and realistic colonoscopy scenarios, we feed video frames of the colonoscopy into COLMAP~\cite{schonberger2016structure} to obtain the camera poses and camera intrinsics. Following the conversion of COLMAP poses from their original camera coordinates to the world coordinate system, we proceed to store this camera trajectory in the TUM format with suitable scaling. Additionally, we will adjust the estimated camera intrinsics to align with the resizing and cropping processes applied to the final reconstructed RGB data.

% \begin{figure}[ht]
% 	\centering
% 	\includegraphics[width=0.95\textwidth]{figure/depth1111.pdf}
%  	\vspace{-2mm}
% 	\caption{Qualitative results of different depth estimation methods. It shows that \textit{ToDER} yields an accurate depth estimation, especially the details indicated within the blue box.}
% 	\label{fig:depth_syn}
%  	\vspace{-2mm}
% \end{figure}

% \begin{figure}[ht]
% 	\centering
% 	\includegraphics[width=0.7\textwidth]{figure/depth22.pdf}
%  	\vspace{-2mm}
% 	\caption{Depth estimation results for realistic colonoscopy. In comparison to other methods, \textit{ToDER} exhibits a higher capability for generating reasonable depths estimations, as indicated by the blue box. This observation underscores its ability to capture geometric details effectively.}
% 	\label{fig:depth_real}
%  	\vspace{-2mm}
% \end{figure}

\begin{table*}[!t]
\caption{The quantitative results of depth estimation. This table demonstrates the superiority of \textit{ToDER} compared with advanced self-supervised and domain adaptation depth estimation methods. Additionally, the effect of the TNet and the bi-directional structure is also analyzed as ablation experiments.}
\resizebox{1.0\textwidth}{!}{
\begin{tabular}{c|cccc|ccc}
\multirow{2}{*}{ Methods } & \multicolumn{4}{c|}{ Error metric $\downarrow$} & \multicolumn{3}{c}{ Accuracy metric $\uparrow$} \\
 & Abs Rel & Sq Rel & RMSE & RMSE log & $\delta<1.25$ & $\delta<1.25^2$ & $\delta<1.25^3$  \\
\hline SfMLearner~\cite{zhou2017unsupervised} & $0.306$ & $1.964$ & $4.369$ & $0.347$ & $0.601$ & $0.804$ & $0.899$  \\
SC-SfMLearner~\cite{bian2019unsupervised} & $0.248$ & $1.643$ & $3.818$ & $0.286$ & $0.674$ & $0.844$ & $0.927$   \\
Monodepth2~\cite{godard2019digging} & $0.164$ & $0.840$ & $3.539$ & $0.239$ & $0.752$ & $0.913$ & $0.965$  \\
S2RDepthNet~\cite{chen2021s2r} & $0.237$ & $1.207$ & $3.700$ & $0.290$ & $0.656$ & $0.887$ & $0.956$  \\

\hline 
  Ours - bi-direct & $0.286$ & $1.379$ & $3.889$ & $1.037$ & $0.554$ & $0.799$ & $0.880$ \\
  Ours - TNet & $0.155$ & $0.434$ & $2.300$ & $0.200$ & $0.786$ & $0.955$ & $0.985$ \\
 Ours & $\textbf{0.144}$ & $\textbf{0.390}$ & $\textbf{2.190}$ & $\textbf{0.184}$ & $\textbf{0.818}$ & $\textbf{0.964}$ & $\textbf{0.987}$ \\
\hline
\end{tabular}
}
\label{table:depth}
\end{table*}

% 这里可以写的详细一些，可以列举公式
% \subsection{Depth Estimation}
% \subsubsection{Evaluation metrics}To assess the performance of depth estimation, Eigen \textit{et al.}~\cite{eigen2014depth} proposed five widely accepted evaluation metrics. The first metric is absolute relative difference (Abs Rel). It mitigates the influence of large errors by normalizing per-pixel difference. The definition of Abs Rel is as follows:

% \begin{equation}
%   Abs Rel =  \frac{1}{N}\sum\frac{|d_i-d_i^*|}{d_i}
% \label{eq:abs rel}
% \end{equation}
% where $d_i$ and $d_i^*$ represent the ground truth and predicted depth at pixel $i$, and $N$ denotes the total number of pixels.

% The second metric is square relative error (Sq Rel). Sq Rel represent the square term penalizes larger depth error. We define the Sq Rel to be:
% \begin{equation}
%   Sq Rel =  \frac{1}{N}\sum\frac{|d_i-d_i^*|^2}{d_i}
% \label{eq:Sq rel}
% \end{equation}

\begin{figure*}[ht]
	\centering
	\includegraphics[width=0.99\textwidth]{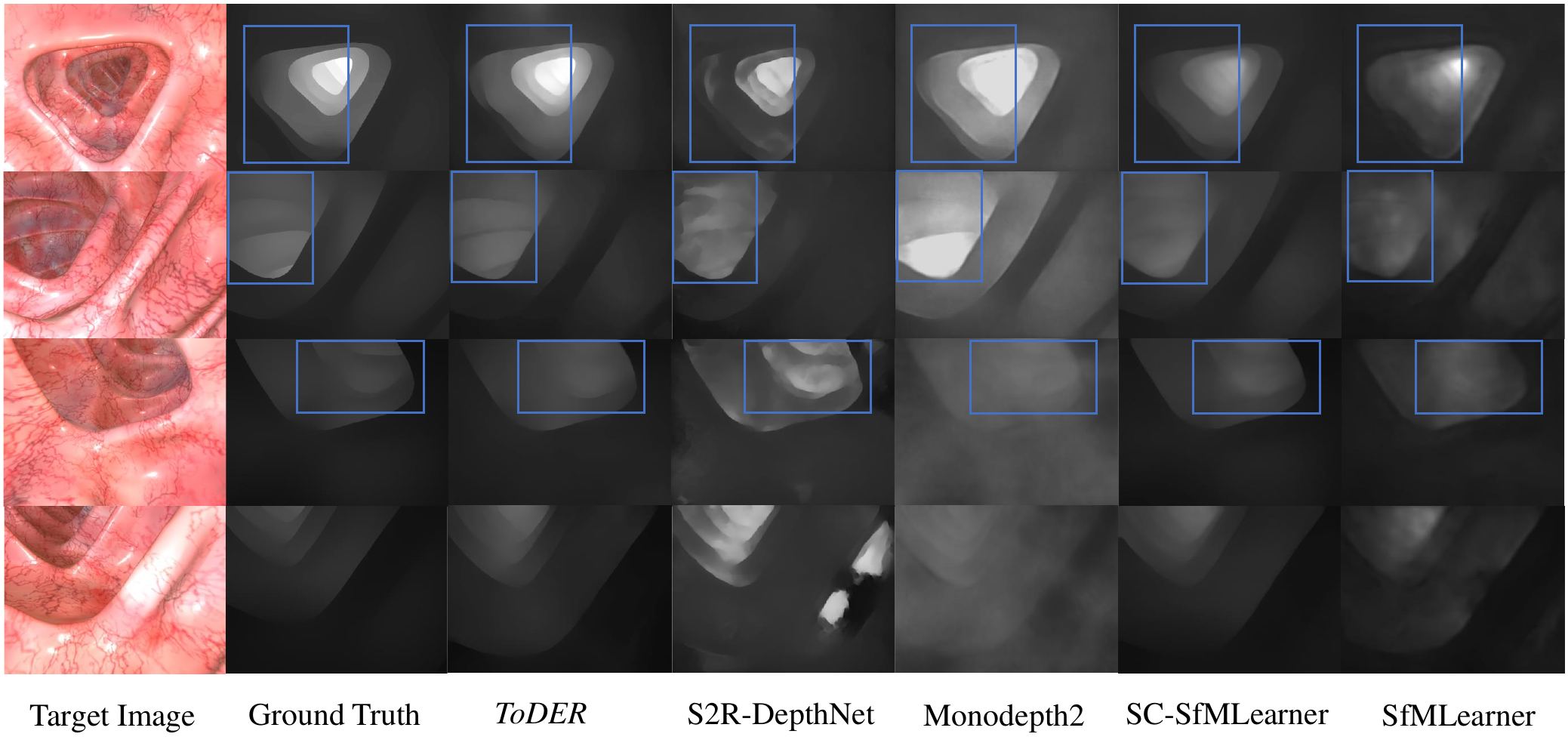}
 	\vspace{-4mm}
	\caption{\textbf{Qualitative results on synthetic videos.} It shows that \textit{ToDER} performs superior depth estimation compared with modern depth estimation methods.}
	\label{fig:depth_synth}
 	\vspace{-0mm}
\end{figure*}

% Then we use root mean square error (RMSE) and $RMSE_{log}$ to evaluate regression errors. They would be defined as:

% \begin{equation}
% \begin{aligned}
% & RMSE =  \sqrt{\frac{1}{N}\sum{|d_i-d_i^*|^2}} \\
% & RMSE_{log} = \sqrt{\frac{1}{N}\sum{|\log_{d_i}-\log_{d_i^*}|^2}}
% \label{eq:RMSE}
% \end{aligned}
% \end{equation}

% At last, we use accuracies with a threshold $(\delta t=1,2,3)$ to evaluate the accuracy of our depth map at the pixel level.

% \begin{figure*}[ht]
% 	\centering
% 	\includegraphics[width=0.78\textwidth]{figure/syn2.pdf}
%  	\vspace{-3mm}
% 	\caption{The reconstructed colons from synthetic data. The results are obtained through depth maps from the ground truth, our \textit{ToDER}, S2R-DepthNet, and Monodepth2. Compared to other approaches, our method showcases superior texture fidelity and continuity(red arrows), along with reduced noise levels(blue arrows).}
% 	\label{fig:synthetic}
%  	\vspace{-4mm}
% \end{figure*}

% \begin{figure*}[ht]
% 	\centering
% 	\includegraphics[width=0.78\textwidth]{figure/Reconstruction.pdf}
%  	\vspace{-3mm}
% 	\caption{The reconstructed colons used for quantitative evaluation. The results are obtained through depth maps from our \textit{ToDER}, S2R-DepthNet and Monodepth2.}
% 	\label{fig:synthetic2}
%  	\vspace{-4mm}
% \end{figure*}
\subsection{Depth Estimation}
\subsubsection{Evaluation on Synthetic Data} 
We conduct quantitative evaluations on synthetic datasets, leveraging available ground truths and  comparing \textit{ToDER} with contemporary self-supervised methods~\cite{godard2019digging,zhou2017unsupervised,bian2019unsupervised} and domain adaptation methods~\cite{chen2021s2r} as shown in Tab.~\ref{table:depth}. \textit{ToDER} outperforms these methods on all metrics, showcasing the effectiveness of our method in depth estimation. We use four test sets each in domains A and B, applying A-to-B and B-to-A models for domain adaptation evaluations and the respective domain models for self-supervised approaches, averaging results across all eight sets. Further qualitative analysis as shown Fig.~\ref{fig:depth_synth} demonstrates the ability of \textit{ToDER} to accurately depict colon depths structure. While other methods provide good estimates, only \textit{ToDER} outputs clear depth structures in the intestine's far end, closely resembling ground truths. Competing methods, such as S2R-Depthnet and Monodepth2, often show inaccuracies in distant areas or incomplete structural representation. The precision and clarity in \textit{ToDER}'s results are attributed to the insights from the bi-directional domain adaptation process with geometric constraints.

\subsubsection{Evaluation on Realistic Data} 
\begin{figure}[h]
	\centering
	\includegraphics[width=0.72\textwidth]{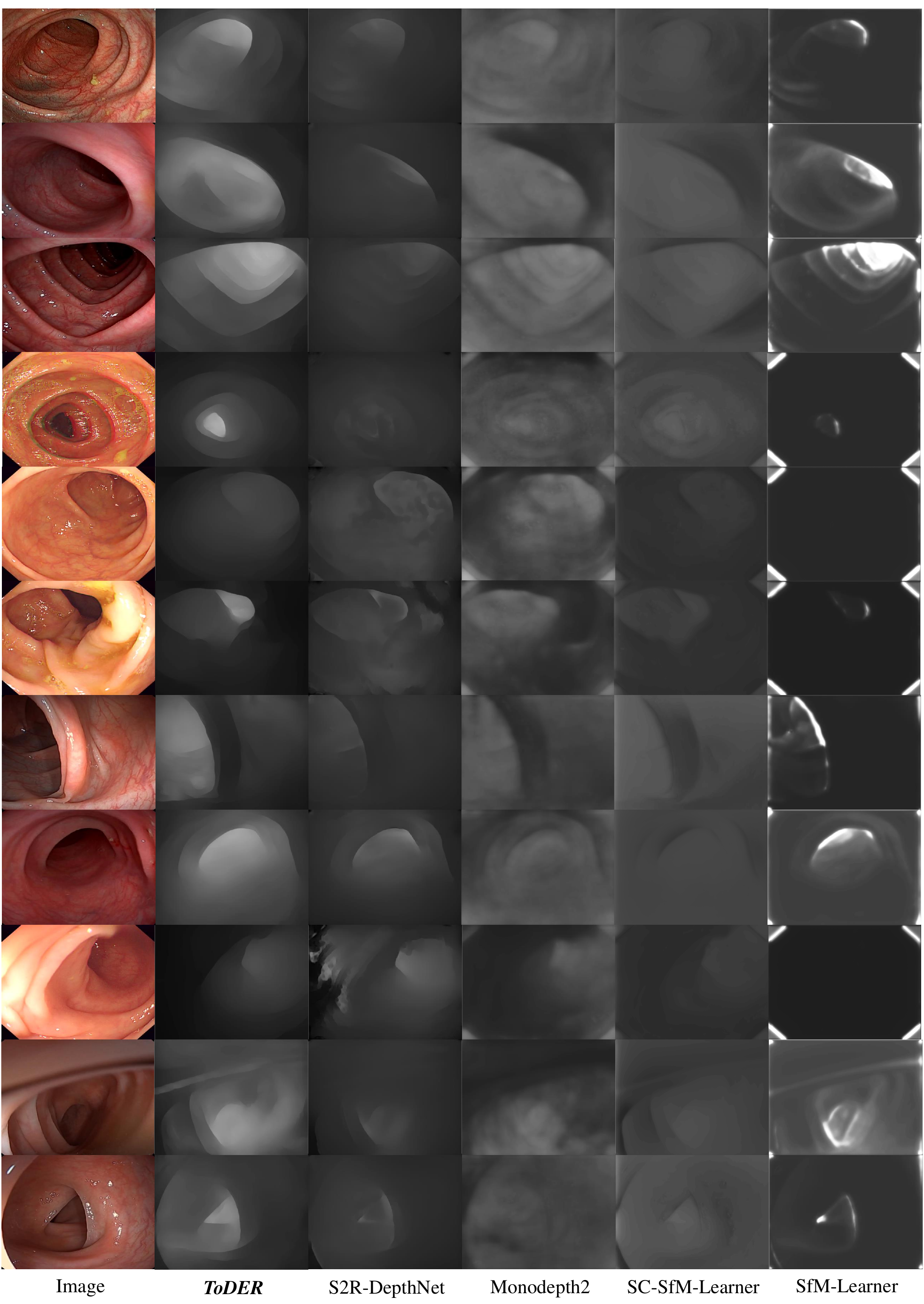}
 	\vspace{-4mm}
	\caption{\textbf{More qualitative depth estimation in realistic colonoscopy videos. Note that the ground truth depths of these realistic colonoscopy videos are not available. We are able to capture the geometry of the colon structure and obtain stable depth maps while other methods produce results with noise and perform relatively unstable prediction.}}
	\label{fig:depth_more}
 	\vspace{-0mm}
\end{figure}

\textit{ToDER} demonstrates satisfactory performance in realistic scenarios, as depicted in Fig.~\ref{fig:depth_real}. We do not have ground truth for evaluation but the depth maps generated by compared methods display similar flaws as they do in the former synthetic case. The depth maps from Monodepth2 exhibit a deficiency in contrast and the edges of colon folds appear unclear, which could cause undesired extra meshes during reconstruction. And as anticipated, S2R-DepthNet generates color patches in the area marked with blue boxes. More qualitative depth estimation results are available in Fig.~\ref{fig:depth_more}.

\subsubsection{Ablation experiment}
We conduct an ablation study as the lower part of Table~\ref{table:depth} shows. ``Ours - bi-direct'' denotes a single directional domain adaptation that only converts the source into the target domain and ``Ours-TNet'' denotes \textit{ToDER} model without TNet. As the table~\ref{table:depth} shows, our full pipeline outperforms other configurations in terms of depth map quality across all evaluation metrics. %This leads us to the conclusion that the to-and-fro structure is a crucial design for depth estimation, and TNet further enhances the performance of our method by introducing epipolar constraints.

\begin{figure*}[t]
	\centering
	\includegraphics[width=1.0\textwidth]{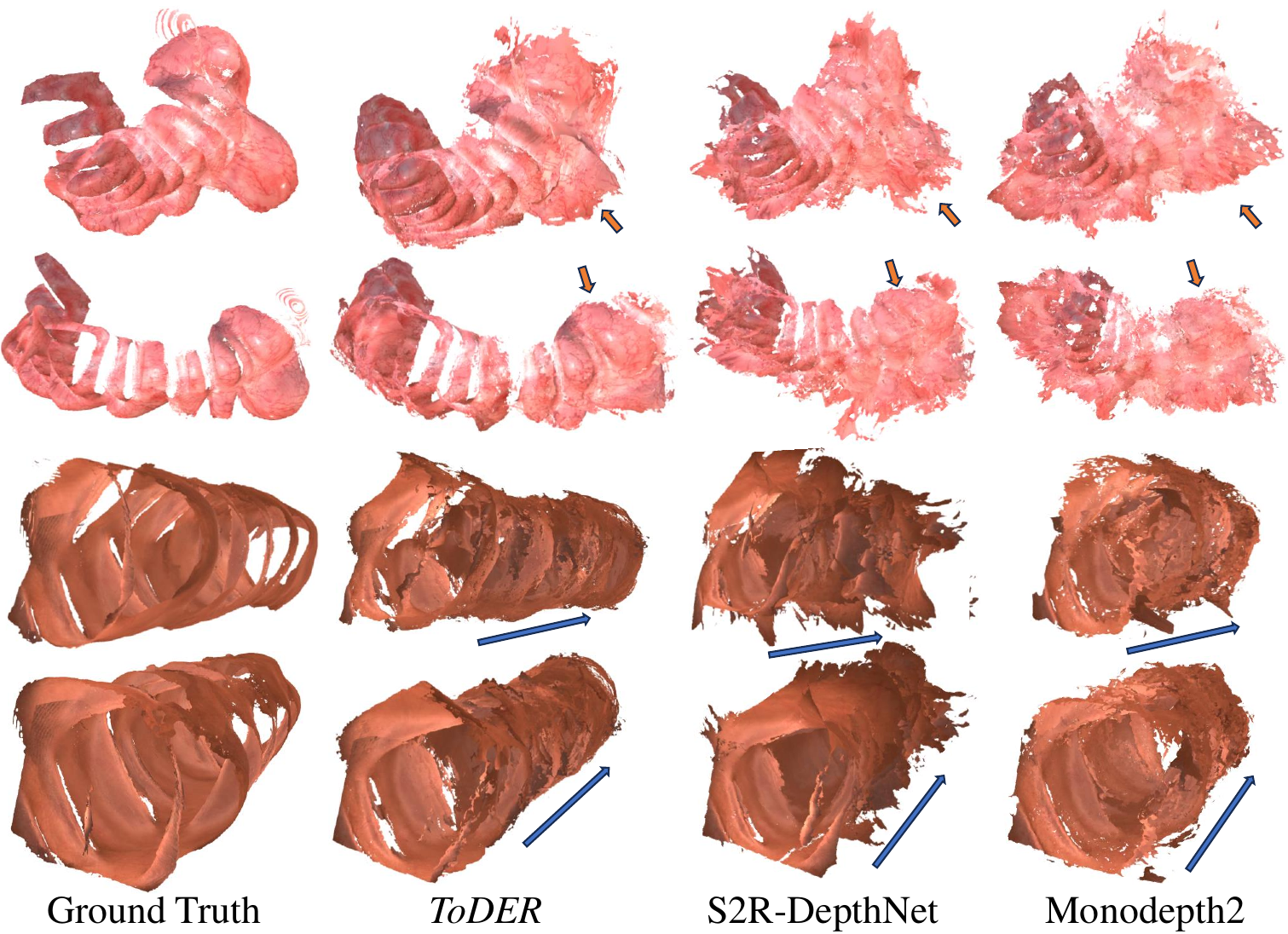}
 	\vspace{-4mm}
	\caption{The reconstructed colons from synthetic data. The results are obtained through depth maps from the ground truth, our \textit{ToDER}, S2R-DepthNet, and Monodepth2. \textit{ToDER} showcases superior texture fidelity and continuity(red arrows), along with reduced noise levels (blue arrows)}
	\label{fig:recon_synth}
 	\vspace{-4mm}
\end{figure*}

\subsection{Reconstruction}

We reconstruct 3D colon models from colonoscopy videos with poses from \cite{schonberger2016structure} and predicted depth maps, as per \cite{schops2019surfelmeshing}. Benefiting from the structural prior knowledge through domain adaptation and the self-supervised constraint, the accurate depth maps predicted for real scenarios allow high-quality reconstructions through the proposed pipeline. COLMAP, due to variations in lighting and inadequate feature points, tends to produce erroneous feature matching, resulting in cluttered and noisy reconstructions. As shown in Fig.~\ref{fig:teaser}, our reconstruction performed better in texture details, with minimal noise, indicating its ability to be applied in real situations. 

To provide a quantitative comparison, we conduct an experiment on our synthetic data. We compute the cloud-mesh distances between the reconstructed results and the ground truth models using CloudCompare~\cite{girardeau2016cloudcompare} after fine registration with Iterative Closest Point algorithm. The computed distances are presented in Tab.~\ref{table:recon} and the corresponding visualization results in Fig.~\ref{fig:recon_synth}. The mean and standard deviation distances between our reconstruction and the ground truth are smaller and our reconstruction exhibits a higher density.

% As shown in Fig.~\ref{fig:recon_synth}, \textit{ToDER} outshines both self-supervised \cite{godard2019digging} and domain adaptation \cite{chen2021s2r} methods in reconstruction quality. Different from S2R-Depthnet's one-way adaptation, \textit{ToDER} uses bi-directional domain adaptation, combining pose estimation and epipolar geometry for enhanced structural accuracy in reconstructions. \textit{ToDER} outperforms Monodepth2 by integrating extensive intestinal structure knowledge, leading to more detailed reconstructions and reduced noise through depth consistency enforcement.

%A lower mean value signifies higher accuracy in the reconstruction, indicating a more precise alignment of the reconstructed model with the ground truth. Similarly, a smaller variance value indicates a reduced presence of significant deviations, thereby presenting less noise within the reconstructed model.

% The reconstruction method we propose has two advantages. Firstly, it takes as input the entire RGB image along with depth information corresponding to each pixel, allowing us to achieve denser reconstructions. Secondly, due to the constraints we impose on the continuity of the depth map, our reconstruction results are less prone to flaws that result in multiple reconstructions of the same surface.

\begin{table}%[t]
\centering
\caption{The quantitative results of 3D reconstruction.}
\resizebox{0.75\textwidth}{!}{
\begin{tabular}{c|cccc}
\multirow{1}{*}{ Methods }  & Ours & S2R-DepthNet & Monodepth2   \\
\hline Mean Distance & $\textbf{0.176}$ & $0.258$ & $0.767$   \\
\hline Standard Deviation & $\textbf{0.163}$ & $0.244$ & $0.744$ 
\\
\hline
\end{tabular}
}
\label{table:recon}
\vspace{-4mm}
\end{table}
\section{Conclusion}
% In this paper, we propose a novel depth estimation and reconstruction pipeline for optical colonoscopy. Our method leverages a bi-directional adaptation architecture and improves the depth guidance from both the realistic and synthetic source domain. Geometry constraints provided by our proposed TNet further improve the depth accuracy. We conduct comprehensive experiments to demonstrate the superiority of our method in depth estimation and reconstruction qualitatively and quantitatively. Depth produced through our method on realistic video displays reasonable details and allows excellent reconstruction results in the final output. Our method boosts the visualization of optical colonoscopy, suggesting possible unobserved regions, and therefore has the potential to mitigate the misdiagnosis rate of colon disease clinically.

In this work, we propose a novel depth estimation and reconstruction pipeline for optical colonoscopy. Our bi-directional adaptation network improves depth guidance from both realistic and synthetic data. Geometry constraints from our proposed TNet boost depth accuracy. Comprehensive experiments show our method outperforms qualitatively and quantitatively in depth estimation and reconstruction. Depths from our method on realistic videos show reasonable details and allow excellent reconstructions. Our pipeline enhances colonoscopy visualization, suggests potential missed regions and could help reduce clinical misdiagnosis of colon diseases.
\bibliographystyle{splncs04}
\bibliography{ref}

\begin{thebibliography}{10}
\providecommand{\url}[1]{\texttt{#1}}
\providecommand{\urlprefix}{URL }
\providecommand{\doi}[1]{https://doi.org/#1}

\bibitem{bian2021unsupervised}
Bian, J.W., Zhan, H., Wang, N., Li, Z., Zhang, L., Shen, C., Cheng, M.M., Reid, I.: Unsupervised scale-consistent depth learning from video. International Journal of Computer Vision  \textbf{129}(9),  2548--2564 (2021)

\bibitem{bian2019unsupervised}
Bian, J., Li, Z., Wang, N., Zhan, H., Shen, C., Cheng, M.M., Reid, I.: Unsupervised scale-consistent depth and ego-motion learning from monocular video. Advances in neural information processing systems  \textbf{32} (2019)

\bibitem{chen2019slam}
Chen, R.J., Bobrow, T.L., Athey, T., Mahmood, F., Durr, N.J.: Slam endoscopy enhanced by adversarial depth prediction. arXiv preprint arXiv:1907.00283  (2019)

\bibitem{chen2021s2r}
Chen, X., Wang, Y., Chen, X., Zeng, W.: S2r-depthnet: Learning a generalizable depth-specific structural representation. In: Proceedings of the IEEE/CVF conference on computer vision and pattern recognition. pp. 3034--3043 (2021)

\bibitem{cordts2016cityscapes}
Cordts, M., Omran, M., Ramos, S., Rehfeld, T., Enzweiler, M., Benenson, R., Franke, U., Roth, S., Schiele, B.: The cityscapes dataset for semantic urban scene understanding. In: Proceedings of the IEEE conference on computer vision and pattern recognition. pp. 3213--3223 (2016)

\bibitem{girardeau2016cloudcompare}
Girardeau-Montaut, D.: Cloudcompare. France: EDF R\&D Telecom ParisTech  \textbf{11} (2016)

\bibitem{godard2019digging}
Godard, C., Mac~Aodha, O., Firman, M., Brostow, G.J.: Digging into self-supervised monocular depth estimation. In: Proceedings of the IEEE/CVF international conference on computer vision. pp. 3828--3838 (2019)

\bibitem{hong2007colonoscopy}
Hong, W., Wang, J., Qiu, F., Kaufman, A., Anderson, J.: Colonoscopy simulation. In: Medical Imaging 2007: Physiology, Function, and Structure from Medical Images. vol.~6511, pp. 212--219. SPIE (2007)

\bibitem{incetan2021vr}
{\.I}ncetan, K., Celik, I.O., Obeid, A., Gokceler, G.I., Ozyoruk, K.B., Almalioglu, Y., Chen, R.J., Mahmood, F., Gilbert, H., Durr, N.J., et~al.: Vr-caps: a virtual environment for capsule endoscopy. Medical image analysis  \textbf{70},  101990 (2021)

\bibitem{kingma2014adam}
Kingma, D.P., Ba, J.: Adam: A method for stochastic optimization. arXiv preprint arXiv:1412.6980  (2014)

\bibitem{li2018megadepth}
Li, Z., Snavely, N.: Megadepth: Learning single-view depth prediction from internet photos. In: Proceedings of the IEEE conference on computer vision and pattern recognition. pp. 2041--2050 (2018)

\bibitem{ma2021rnnslam}
Ma, R., Wang, R., Zhang, Y., Pizer, S., McGill, S.K., Rosenman, J., Frahm, J.M.: Rnnslam: Reconstructing the 3d colon to visualize missing regions during a colonoscopy. Medical image analysis  \textbf{72},  102100 (2021)

\bibitem{mahmood2018unsupervised}
Mahmood, F., Chen, R., Durr, N.J.: Unsupervised reverse domain adaptation for synthetic medical images via adversarial training. IEEE transactions on medical imaging  \textbf{37}(12),  2572--2581 (2018)

\bibitem{mur2015orb}
Mur-Artal, R., Montiel, J.M.M., Tardos, J.D.: Orb-slam: a versatile and accurate monocular slam system. IEEE transactions on robotics  \textbf{31}(5),  1147--1163 (2015)

\bibitem{posner2022c}
Posner, E., Zholkover, A., Frank, N., Bouhnik, M.: C 3 fusion: consistent contrastive colon fusion, towards deep slam in colonoscopy. In: International Workshop on Shape in Medical Imaging. pp. 15--34. Springer (2023)

\bibitem{rau2019implicit}
Rau, A., Edwards, P.E., Ahmad, O.F., Riordan, P., Janatka, M., Lovat, L.B., Stoyanov, D.: Implicit domain adaptation with conditional generative adversarial networks for depth prediction in endoscopy. International journal of computer assisted radiology and surgery  \textbf{14},  1167--1176 (2019)

\bibitem{schonberger2016structure}
Schonberger, J.L., Frahm, J.M.: Structure-from-motion revisited. In: Proceedings of the IEEE conference on computer vision and pattern recognition. pp. 4104--4113 (2016)

\bibitem{schops2019surfelmeshing}
Sch{\"o}ps, T., Sattler, T., Pollefeys, M.: Surfelmeshing: Online surfel-based mesh reconstruction. IEEE transactions on pattern analysis and machine intelligence  \textbf{42}(10),  2494--2507 (2019)

\bibitem{sturm12iros}
Sturm, J., Engelhard, N., Endres, F., Burgard, W., Cremers, D.: A benchmark for the evaluation of rgb-d slam systems. In: Proc. of the International Conference on Intelligent Robot Systems (IROS) (Oct 2012)

\bibitem{zhang2020template}
Zhang, S., Zhao, L., Huang, S., Ye, M., Hao, Q.: A template-based 3d reconstruction of colon structures and textures from stereo colonoscopic images. IEEE Transactions on Medical Robotics and Bionics  \textbf{3}(1),  85--95 (2020)

\bibitem{zhang2021colde}
Zhang, Y., Frahm, J.M., Ehrenstein, S., McGill, S.K., Rosenman, J.G., Wang, S., Pizer, S.M.: Colde: a depth estimation framework for colonoscopy reconstruction. arXiv preprint arXiv:2111.10371  (2021)

\bibitem{zheng2018t2net}
Zheng, C., Cham, T.J., Cai, J.: T2net: Synthetic-to-realistic translation for solving single-image depth estimation tasks. In: Proceedings of the European conference on computer vision (ECCV). pp. 767--783 (2018)

\bibitem{Zheng_2018_ECCV}
Zheng, C., Cham, T.J., Cai, J.: T2net: Synthetic-to-realistic translation for solving single-image depth estimation tasks. In: Proceedings of the European Conference on Computer Vision (ECCV) (September 2018)

\bibitem{zhou2017unsupervised}
Zhou, T., Brown, M., Snavely, N., Lowe, D.G.: Unsupervised learning of depth and ego-motion from video. In: Proceedings of the IEEE conference on computer vision and pattern recognition. pp. 1851--1858 (2017)

\bibitem{zhu2017unpaired}
Zhu, J.Y., Park, T., Isola, P., Efros, A.A.: Unpaired image-to-image translation using cycle-consistent adversarial networks. In: Proceedings of the IEEE international conference on computer vision. pp. 2223--2232 (2017)

\end{thebibliography}

\clearpage
\section{Appendix}

\end{document}